\documentclass{article}
\usepackage[frozencache=true,cachedir=minted-cache]{minted} 
\usepackage[dvipsnames]{xcolor}
\definecolor{bg}{rgb}{0.95,0.95,0.95}
\setminted[python]{bgcolor=bg}

\usepackage{caption} 
\newcommand\FlexModel{\texttt{FlexModel}}
\newcommand\HookFunction{\texttt{HookFunction}}

\pdfoutput=1

\PassOptionsToPackage{numbers}{natbib}

\usepackage[final]{neurips_2023}
\usepackage{graphicx}
\usepackage{subcaption}
\usepackage{booktabs}

\usepackage[utf8]{inputenc} 
\usepackage[T1]{fontenc}    
\usepackage{hyperref}       
\usepackage{url}            
\usepackage{booktabs}       
\usepackage{amsfonts}       
\usepackage{nicefrac}       
\usepackage{microtype}      
\usepackage{xcolor}         
\usepackage{amsmath}
\hypersetup{
    colorlinks=true,
    urlcolor=cyan,
    }

\title{FlexModel: A Framework for Interpretability of Distributed Large Language Models}

\author{%
  Matthew Choi\thanks{Correspondence to \texttt{\{matthew.choi, adil.asif\}@vectorinstitute.ai}}, ~Muhammad Adil Asif,$^*$ John Willes, David B. Emerson\\
  Vector Institute, Toronto, ON, Canada
}

\begin{document}

\maketitle

\begin{abstract}
With the growth of large language models, now incorporating billions of parameters, the hardware prerequisites for their training and deployment have seen a corresponding increase. Although existing tools facilitate model parallelization and distributed training, deeper model interactions, crucial for interpretability and responsible AI techniques, still demand thorough knowledge of distributed computing. This often hinders contributions from researchers with machine learning expertise but limited distributed computing background. Addressing this challenge, we present \FlexModel{}, a software package providing a streamlined interface for engaging with models distributed across multi-GPU and multi-node configurations. The library is compatible with existing model distribution libraries and encapsulates PyTorch models. It exposes user-registerable \HookFunction{}s to facilitate straightforward interaction with distributed model internals, bridging the gap between distributed and single-device model paradigms. Primarily, \FlexModel{} enhances accessibility by democratizing model interactions and promotes more inclusive research in the domain of large-scale neural networks. The package is found at \url{https://github.com/VectorInstitute/flex_model}.
\end{abstract}

\section{Introduction}

Driven by recent advances in language modelling, neural network sizes have increased dramatically \cite{kaplan2020scaling, shoeybi2019megatron}. Large language models (LLMs), with billions of parameters, have demonstrated impressive and surprising abilities, often attributed to their generalization and in-context learning capacity \cite{olsson2022incontext, garg2022can, Brown1}. Increased model size has resulted in a commensurate increase in hardware prerequisites for training and deployment of these models, frequently requiring distributed infrastructure. To meet this need, many technologies have emerged \cite{shoeybi2019megatron, zhao2023pytorch, deep_speed_zero}. While these tools abstract numerous challenges associated with model parallelization and distributed training, non-standard interactions and deeper model manipulations, such as intermediate activation retrieval or model editing, still necessitate a comprehensive understanding of distributed computing. However, such interactions are important elements of many techniques in interpretability and responsible AI \cite{elhage2022solu, nanda_grokking, belrose2023eliciting, elhage2022superposition}. For researchers lacking foundational experience in distributed computing, this represents a significant barrier. As such, a large cohort of researchers find themselves at a disadvantage or unable to contribute, despite possessing valuable machine learning expertise, due to the technical complexities inherent in working with distributed models.

In response to this challenge, this paper introduces \FlexModel{}, a software package engineered to deliver a lightweight and user-friendly interface for interacting with large-scale models deployed in multi-GPU and multi-node settings. The library is designed to seamlessly integrate with existing technologies employed by researchers and practitioners, ensuring compatibility and convenience. \FlexModel{} wraps PyTorch models deployed using any of the major libraries, including Accelerate, Fully Sharded Data Parallel (FSDP), and DeepSpeed, and supports user-defined \HookFunction{}s to enable straightforward interaction with distributed model internals during both forward and backward passes. These \HookFunction{}s abstract the required distributed communication and allow researchers to interact with distributed models as if they were running on a single device. 

The contributions of this work are summarized as:
\begin{itemize}
    \item \FlexModel{}: A software package which provides an infrastructure-agnostic model interface to enable researchers to perform interpretability and responsible AI research at scale without a deep understanding of distributed systems. The package aligns distributed model interaction with the simpler paradigm of single-device model manipulation.
    \item Validation of \FlexModel{} for use at scale in two experimental settings: Transformer Induction Head Isolation \cite{olsson2022incontext} and a TunedLens \cite{belrose2023eliciting} implementation.
\end{itemize}

\section{Background and Related Work}

\subsection{Interpretability}

With the release of highly performant, closed-source LLMs, such as ChatGPT and BARD, interest in LLMs has rapidly increased. Open-source models, including LLaMA \cite{touvron2023llama, touvron2023llama2}, have also gained widespread adoption. The rise of LLMs has also attracted attention from researchers interested in model interpretability and explainability. Characterizing the emergence of capabilities in LLMs during training is crucial to understanding their formation and function in downstream applications \cite{bommasani2022opportunities}. Tracing inference and investigating the influence of context are also key to providing insights into model behaviour. However, such investigations are challenging due to the massive number of parameters and layers of non-linearity present in LLMs. 

Typical interpretability and explainability work investigates LLMs on a macroscopic level, where qualitative behaviours are explored. Many such works leverage prompts for such investigation \cite{liang2022holistic, parrish-etal-2022-bbq, kojima2023large, jung-etal-2022-maieutic, wei2023larger}. A more limited set of works are beginning to delve into the internal mechanics of LLMs through probing mechanisms like activation or prompt tuning \cite{azaria2023internal, tian2023softprompt}. Alternatively, mechanistic interpretability develops hypotheses about basic transformer behaviour, derived from first principles. Such approaches aim to more deeply investigate the fundamental components driving LLM generation and inject or train modules to examine the internal representations of these models.

\subsection{Distributed Models and Parallelism} \label{distributed_models_parallelism}

Running inference on models with tens or hundreds of billions of parameters often requires multi-GPU coordination to fit the model parameters in GPU DRAM. Training such models requires multi-node techniques to compute and store activations, gradients and optimizer states. Typically these techniques involve splitting or replicating a model along three ``dimensions,'' namely the data parallel (DP), tensor parallel (TP) and pipeline parallel (PP) dimensions (see Figure \ref{fig:parallel3d}). DP methods, like PyTorch's \texttt{DistributedDataParallel} \cite{liddp2020}, replicate a model on each GPU, effectively increasing the global batch size. Note that the model memory footprint remains the same on each GPU in this case. Frameworks such as \texttt{FullyShardedDataParallel} \cite{zhao2023pytorch} implement sharded DP to address this limitation, where each GPU only holds a piece of the model states. Computation is performed by sequentially communicating these pieces across all GPUs. Alternatively, TP and PP schemes directly seek to reduce the memory footprint of model states. These schemes shard a model instance across multiple GPUs. TP  breaks layer operations into smaller parallel operations distributed over multiple GPUs, but requires expensive all-to-all communication \cite{shoeybi2019megatron}. PP splits the model into sequential stages where forward and backward passes iterate through each stage \cite{huang2019gpipe,shoeybi2019megatron,narayanan2021efficient}. The sequential pipeline stages reduce the per-device memory footprint at the cost of increasing GPU idle time due to data-dependencies, known as pipeline bubbles.

\subsection{Related Work}

Several libraries facilitating model inspection, probing, and interpretability exist. However, each have drawbacks, especially in distributed settings, which \FlexModel{} aims to address. TransformerLens \cite{nandatransformerlens2022} provides a variety of features for interpretability research on transformers, such as activation retrieval and editing. However, models must be re-engineered to fit specific requirements, and features for experimentation in distributed settings are limited. CircuitsVis \cite{CircuitsVis} is a dynamic library for visualizing transformer mechanics, but it assumes access to the target quantities rather than providing probing tools. Finally, Microscope \cite{microscope} and Neuroscope \cite{neuroscope} focus on studying the formation of features and tracing activations in deep learning models. Microscope solely considers vision models, is not publicly available, and only a fixed set of models are admissible. Neuroscope is capable of identifying maximally activating examples for each neuron in several language models. However, the set of supported models is limited with the largest having only 1.4B parameters. 

\section{FlexModel}

Working with LLMs typically involves many painful lessons in distributed systems. Researchers often need to learn the PyTorch distributed back-end and familiarize themselves with many of the distributed model strategies described in Section \ref{distributed_models_parallelism}. Performing interpretability research in such settings, where model surgery and inspection requires opening the black box, is even more challenging.
\FlexModel{} lowers these barriers to interpretability research in LLMs and beyond. The design goals of the \FlexModel{} API are two-fold:
\begin{enumerate}
    \item \textbf{Intuitive:} Applying the \FlexModel{} wrapper to a PyTorch \texttt{nn.Module} should simply add features for model inspection to the target model. Unwrapping the model should produce the original model without side-effects. The \HookFunction{}'s editing function should allow arbitrary code to be run on the activations.
    \item \textbf{Scalable:} \FlexModel{} should be agnostic to the number of GPUs or GPU nodes, the model architecture (e.g. LLaMA, Falcon, GPT), the model size, and the distribution strategy (e.g. DP, FSDP, TP, PP) or composition thereof. 
\end{enumerate}

\subsection{API}

\subsubsection{\FlexModel{}}

The \FlexModel{} class provides the main interface for user interactions. It inherits from \texttt{nn.Module}, allowing developers to easily interact with the wrapped model via the \texttt{nn.Module} API without any code changes. This also enables \FlexModel{} to intercept API calls and inject additional logic if required. \FlexModel{} requires a few initialization arguments: The \texttt{nn.Module} to wrap, a dictionary object where the \HookFunction{}s send retrieved activations, and the distributed strategy used to launch the wrapped module, defined in terms of TP, PP and DP sizes. For more detail on the use of the distributed strategy information, see Appendix \ref{communcation_system}. See Figure \ref{fig:model-init-short} for an example of instantiating a \FlexModel{} using HuggingFace Accelerate for model distribution. For more detail, see Appendix \ref{sec:flexmodelapi}.

\begin{figure}[ht!]
    \begin{minted}[breaklines, breakindent=7pt, breaksymbolleft=]{python}
    model = AutoModelForCausalLM.from_pretrained("model-name")
    model = accelerator.prepare(model)
    
    output_dict: Dict[str, Tensor] = {}
    model = FlexModel(model, output_dict, data_parallel_size=accelerator.num_processes)
    \end{minted}
    \caption{\FlexModel{} initialization example.}
    \label{fig:model-init-short}
\end{figure}

\subsubsection{\texttt{HookFunction}}

Once a \FlexModel{} has been instantiated, users may define a collection of \HookFunction{}s to perform activation retrieval and/or editing. To create a \HookFunction{}, the user provides the name of the target module/tensor, the expected shape of the activation tensor, and, optionally, an editing function to run, see Figure \ref{fig:hook-func} in Appendix \ref{sec:flexmodelapi}. The expected shape is a tuple of integers representing the shape of the target activation tensor. The user need not fill in all of the dimensions; only the sharded dimensions are required. Other dimensions may simply be annotated as \texttt{None} and are inferred by the communication system. The editing function is defined by the user and must adhere to a specific signature. However, the contents of the function are completely up to the user. The guarantee within the editing function body is single-threaded behaviour and the full activation tensor as an input. There are also additional input options to support other \FlexModel{} features like save contexts and trainable modules. See Appendix \ref{communcation_system}, and Figure \ref{fig:dist-comm} therein, for details on how \FlexModel{} and \HookFunction{} implement these features.

\section{Experiments}

\subsection{Induction Head Isolation}

Introduced in \cite{elhage2021mathematical}, induction heads are a specific type of attention head within transformer models that emerge during pretraining. Olsson et al. \cite{olsson2022incontext} theorize that induction heads are the primary driver for in-context learning in LLMs. An attention head is classified as an induction head if it performs both prefix-matching and copying. Concretely, an induction head performs the operation \texttt{[A][B]$\ldots$[A]$\rightarrow$[B]}, where \texttt{[A]} and \texttt{[B]} are tokens in a sequence. Prefix-matching means that the induction head looks back into the context and attends to the token \texttt{[B]} which was prefixed by \texttt{[A]}. The induction head then copies this pattern by increasing the output logit for token \texttt{[B]}. Induction heads are potentially powerful because they have also been observed performing this algorithm on higher-level concepts \cite{olsson2022incontext}, rather than just token level behaviour.

\begin{figure}[!t]
     \centering
     \begin{subfigure}[b]{0.44\textwidth}
         \centering
         \includegraphics[width=\textwidth]{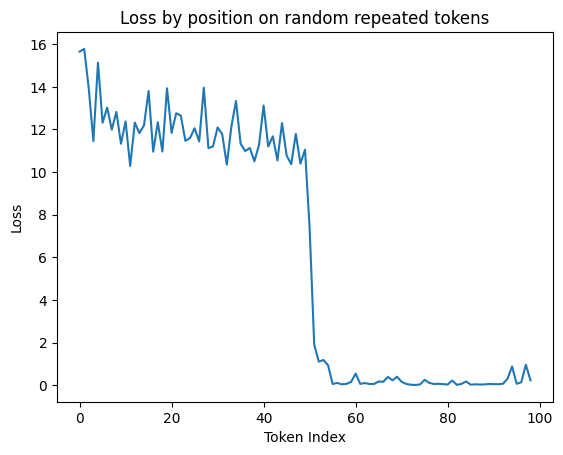}
         \caption{}
         \label{fig:tloss}
     \end{subfigure}
     \begin{subfigure}[b]{0.35\textwidth}
         \centering
         \includegraphics[width=\textwidth]{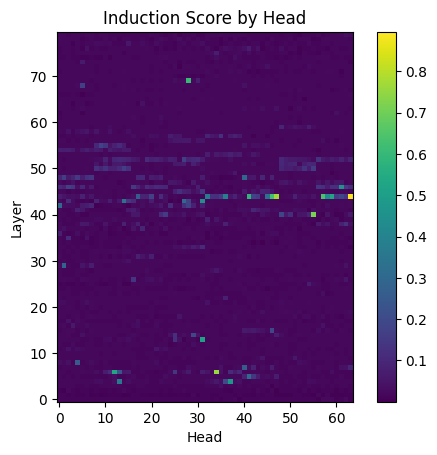}
         \caption{}
         \label{fig:iscore}
     \end{subfigure}
    \caption{Empirical verification of the presence of induction heads within LLaMA-2-70B using a repeated randomly generated sequence as input. The measured loss per token is shown in (\subref{fig:tloss}) and the location of the induction heads is shown in (\subref{fig:iscore}), indicated by the high induction score values.}
\end{figure}

To demonstrate the capabilities of \FlexModel{}, an induction head search within LLaMA-2-70B, distributed using FSDP over four A100 GPUs, is conducted. Results for LLaMA-2-13B are shown in Figure \ref{13b} of Appendix \ref{llama_induction_appendix}. The goal is to identify and isolate any induction heads in the model by retrieving the attention maps of each attention head and evaluating their induction head score. We first uniformly sample 50 tokens, with replacement, to produce a sequence which is repeated once to obtain a 100 token query. Because the first half of the sequence is randomly sampled, the model is expected to poorly predict that portion of the sequence. However, in the presence of induction heads, the model should be able to predict the latter half of the sequence, conditioned on the first half, nearly perfectly, as this aligns with induction head behaviour. The per-token-loss for such a sequence structure is visualized in Figure \ref{fig:tloss}. For a given model head, the induction score is computed by averaging the attention scores $50 - 1 = 49$ tokens back, since induction heads attend to the token \textit{after} the last occurrence of the \textit{current} token, for each sequence position. These scores, across all layers and heads, are shown in Figure \ref{fig:iscore} as a heat map. We observe candidates in the middle layers.

\subsection{TunedLens}
TunedLens \cite{belrose2023eliciting} is an interpretability technique for investigating the residual stream of a transformer model using linear probes \cite{alain2018understanding}. The probes map the transformer layer residual streams to a similar space as the transformer model output residual stream using an affine projection. Thereafter, the final normalization and output unembedding layers are applied, producing token distributions at every model layer. This allows researchers to observe how predictions are iteratively refined during a model's forward pass. To demonstrate the features of \FlexModel{}, a toy experiment implementing TunedLens applied to LLaMA-2-13B is constructed. See Appendix \ref{tunedlens_appendix} for details about the \FlexModel{} implementation of TunedLens, as well as an inference example. Using \FlexModel{} to implement TunedLens greatly reduces the amount of code and decouples TunedLens from the base model.

\subsection{Communication Overhead}
It is important to note that \HookFunction{}s may introduce substantial overhead into model execution, the sources of which stem from editing function compute, added collective communication and device-to-host data movement. Forward-pass editing functions are fully-exposed on the critical path, so latency costs related to host-device synchronization and added compute cannot be hidden. \HookFunction{}s applied to modules with GPU-sharded outputs also incur latency costs, primarily due to added collective communication used to materialize the output activation tensor. Finally, CPU offloading of activation tensors during the execution of many \HookFunction{}s experience the heaviest overhead, due to slow device-to-host data transfers of activation tensors.

To investigate the impact of both the added communication and host memory allocation, we use the PyTorch profiler to measure the execution time of several different experiments. A simulated model is used, comprised of alternating Megatron-LM \texttt{ColumnParallelLinear} and \texttt{RowParallelLinear} layers with ReLU activation functions. In total, the model has $32$ linear layers with a model dimension of $4096$. The model is distributed along the tensor parallel dimension only. The profiles are run on four A100-80G GPUs, with NVLink-v4 intraconnect. The primary metric measured is the time per profiler step, which is averaged over $10$ iterations, as we observed little variance. The time per profile step is calculated as the maximum of the reported \texttt{self\_cuda\_time\_total} and \texttt{self\_cpu\_time\_total}. Refer to Table \ref{fig:profile_results} for the timing results, along with experiment parameters. 

The results of the experiment profiling showed a reasonable increase in latency when extra collective communication operations are introduced. When opting for CPU offloading in favour of GPU offloading, there was a substantial increase in latency. Hence, it is extremely important for \FlexModel{} to avoid frequent CPU offloads when necessary. Using CPU pinned memory mitigates a portion of this overhead. Further discussion on key optimization factors is found in Appendix \ref{communication_overhead_appendix}.

\begin{table}[!t]
\caption{Forward pass latency comparison on a simulated model, with alternating \texttt{ColumnParallelLinear} and \texttt{RowParallelLinear} layers from the Megatron-LM codebase, and a model dimension of 4096. The profiles are run on 4 A100-80G GPUs, and \HookFunction{}s are applied on each linear layer. ``No Hooks'' denotes a regular forward pass. The \FlexModel{} runs vary in distributed communication and local data movement parameters.}
\label{fig:profile_results}
\centering
\begin{tabular}{ccccc}
 & Extra Comm. & Local Data Transfer & Pinned Mem. & Time Per Step (s) \\ 
 \midrule
No Hooks & & & N/A & 0.1237 \\
\FlexModel{} & \checkmark & & N/A & 0.4016 \\
\FlexModel{} & \checkmark & \checkmark & & 6.071 \\
\FlexModel{} & \checkmark & \checkmark & \checkmark & 2.632
\end{tabular}
\end{table}


\section{Conclusion}

In this paper, we introduced a new unified library to advance alignment and interpretability research on LLMs and beyond. Prior works \cite{nandatransformerlens2022, belrose2023eliciting} have successfully provided many of the features that we implement. However, existing works have limited or no support for distributed models, and most libraries only consider a restricted set of compatible models. As such, \FlexModel{} provides a straightforward user interface and an infrastructure-agnostic wrapper around PyTorch models requiring minimal additional code. Additionally, we have shown how \FlexModel{} provides utility for workloads across a selection of current research directions within LLM alignment and interpretability. These include induction head identification, linear probing, activating editing, activation caching, and simple insertion of generic trainable modules. We aim to provide on-going support for \FlexModel{} and implement additional features in future work.

\section*{Social Impacts Statement}

The \FlexModel{} package is designed to make interpretability and explainability research simpler and more intuitive for large models that require distributed compute. Large models, especially in the form of LLMs, are becoming increasingly common and are widely deployed in important settings. Facilitating such research is paramount to ensuring that such models are safe, free from bias, and robust in the wider settings in which they are used.

\bibliography{cites}
\bibliographystyle{plain}

\appendix 
\newpage

\section*{Appendix}

\section{\FlexModel{} Usage} \label{sec:flexmodelapi}

Below, in Figures \ref{fig:model-init} and \ref{fig:hook-func}, are code examples demonstrating how to instantiate a \FlexModel{} object and how to define a custom editing function in the proposed framework, respectively.

\begin{figure}[ht!]
    \begin{minted}[breaklines, breakindent=7pt, breaksymbolleft=]{python}
    from flex_model import FlexModel, HookFunction
    
    accelerator = Accelerator()
    model = AutoModelForCausalLM.from_pretrained("meta-llama/Llama-2-13b-chat-hf")
    tokenizer = AutoTokenizer.from_pretrained("meta-llama/Llama-2-13b-chat-hf")
    
    model = accelerator.prepare(model)
    
    activation_dict: Dist[str, Tensor] = {}
    model = FlexModel(model, activation_dict, data_parallel_size=accelerator.num_processes)
    \end{minted}
    \caption{\FlexModel{} initialization example.}
    \label{fig:model-init}
\end{figure}

\begin{figure}[ht!]
    \begin{minted}[breaklines, breakindent=7pt, breaksymbolleft=]{python}
    def my_edit_fn(current_module, inputs, save_ctx, global_modules) -> Tensor:
        # Cache data for later.
        _, s, _ = torch.svd(inputs)
        save_ctx.activation_singular_values = s
        # Edit the activation tensor.
        inputs = torch.where(inputs > 1.0, inputs, 0.0)
        # Apply a torch layer to the activation tensor.
        outputs = global_modules["linear_projection"](inputs)
        return outputs
    
    my_hook_function = HookFunction(
        "model.layers.23",
        expected_shape=(4, 1024, 4096),
        editing_function=my_edit_fn)
    flex_model.register_hook_function(my_hook_function)
    \end{minted}
    \caption{\HookFunction{} registration example.}
    \label{fig:hook-func}
\end{figure}

\section{\FlexModel{} Communication} \label{communcation_system}

The core design of the \HookFunction{} editing function runtime is single-threaded execution and access to the target unsharded activation tensor for manipulation or storage. Single-threaded execution means user code is run once across all workers. Hence, they need not worry about multi-worker coordination. This is much easier for the user to develop and iterate on, as code running on their local python interpreter is directly transferable to the editing function. The user-defined code is run on the desired unsharded activation tensor. For example, consider a model that is duplicated across 4 GPUs (i.e. DP = 4). The communication system gathers the activations from each duplicated model instance and concatenates them, in the batch dimension, to form the relevant unsharded activation tensor. Additionally, the activation tensors, which are streamed to the \FlexModel{}'s output dictionary, are only present on the rank 0 worker's CPU to prevent additional GPU memory usage.

To enable this functionality, the \FlexModel{} and \HookFunction{} instances coordinate the distributed communication. \FlexModel{} initialize the communication system and provides a pointer to the output dictionary to each registered \HookFunction{} instance. The \HookFunction{}s handle the necessary distributed communication for unsharding of the activation tensor, running the editing function and dispersing the subsequent edited activation tensor.

\begin{figure}[ht!]
     \centering
     \begin{subfigure}[c]{0.33\textwidth}
         \centering
         \includegraphics[width=\textwidth]{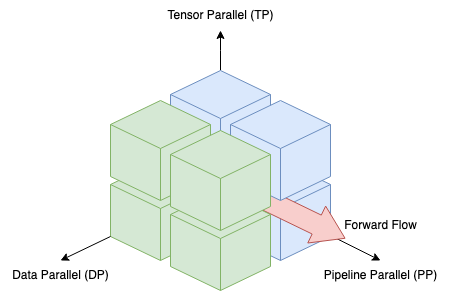}
         \caption{}
         \label{fig:parallel3d}
     \end{subfigure}
     \begin{subfigure}[c]{0.66\textwidth}
         \centering
         \includegraphics[width=\textwidth]{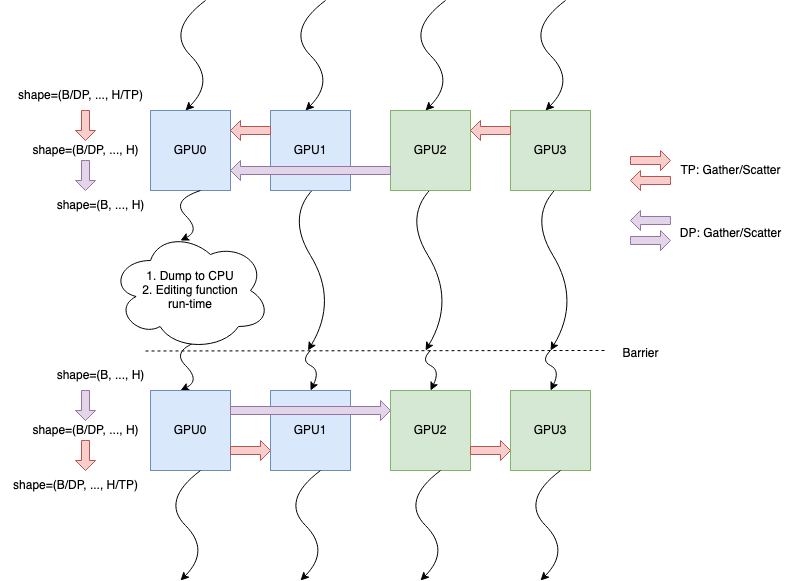}
         \caption{}
         \label{fig:singlethread}
     \end{subfigure}
    \caption{(\subref{fig:parallel3d}) Models distributed over 8 GPUs, depicted as a 3D mesh, have activation tensors distributed along a 2D slice in the TP and DP axes. The process facilitating single-threaded editing is shown in (\subref{fig:singlethread}). Gather collectives are used to unshard the TP then DP axes. The local rank 0 GPU puts a copy of the full activation tensor in CPU and runs the user-provided editing function on it. Finally scatter collectives are used to redistribute shards of the modified tensor first across DP then TP dimensions.
    }
    \label{fig:dist-comm}
\end{figure}

Initialization of the communication system consists of constructing a 3D GPU device mesh, exactly representing the 3D layout of the wrapped model (i.e. DP, TP and PP; see Figure \ref{fig:parallel3d}). PyTorch distributed groups are created such that activation tensors can be gathered, scattered, and singleton accumulated on CPU. Note that gathering and scattering activation tensors only require a subset of the distributed groups and communication collectives used for accumulating activation tensors to the rank 0 CPU. 

To run the single-threaded editing function on the unsharded activation tensor, each \HookFunction{} runs prologue and epilogue functions to support the editing function. The prologue consists of extracting the local activation tensor from the layer outputs of the current module and gathering them along the necessary dimensions to form the full activation tensor. The epilogue is an exact inverse of the prologue wherein it scatters the edited activation tensor along the same dimensions in reverse-order and repacks the edited local activation tensor into the layer outputs. Gathering and scattering are performed along the DP and TP axes, but not the PP axis, as activations are never sharded between different PP stages. Refer to Figure \ref{fig:singlethread} for the full workflow. Since the layer outputs can be arbitrary Python objects, \HookFunction{} uses a tree-traversal strategy to unpack and repack the activation tensors with the layer outputs. These are similar to JAX \texttt{tree\_util} functions \cite{jax2018github}. Finally, activation tensors present on each pipeline stage are gathered to the global rank 0 GPU, and placed onto its CPU.

\section{Communication Overhead\label{communication_overhead_appendix}}
The communication overhead results in Table \ref{fig:profile_results} show a significant increase in latency as collective communication and CPU offloading were added. The baseline "No Hooks" case involved running a standard forward pass through the simulated model. The rest of the cases use \FlexModel{} and complexities are incrementally added, such as extra collective communication to materialize the full activation tensors, local data transfers from device to host and the usage of pinned memory to stage the device-to-host transfers. Since half of the layers in the test model required collective communication (i.e. the \texttt{ColumnParallelLinear} layers), there are $16$ additional \texttt{all-gather} collective communication operations performed. This results in \textasciitilde 3x additional latency. Note that the GPU offload itself does not result in any significant added overhead since the activation tensor is already resident in device memory. The bulk of the overhead stems from the CPU offloading operation, which occurs at every layer in the model. Offloading to pinned memory before pageable memory is much faster than naively offloading immediately to pageable memory, at 6.5x increased latency compared to 15x increased latency relative to GPU offloading. 

After profiling many \HookFunction{} scenarios, it is immediately obvious that mitigating overhead from copy operations is of the utmost importance. Currently, activation tensors are guaranteed to be offloaded to the global rank 0 GPU DRAM (or local CPU RAM). However, for frameworks such as TunedLens, this results in an indirection where \FlexModel{} offloads tensors to rank 0, but the tensors must be moved again to where the TunedLens model lives. This is further exacerbated by the NCCL requirement for tensors to be resident in device memory, which incurs a host-to-device transfer if activation tensors were offloaded to host memory. Therefore an obvious optimization would be to achieve zero-copy activation tensor routing, where the user may specify the required offload location of tensors or groups of tensors. Such a strategy avoids unnecessary memory allocations and/or copying operations. 

\section{TunedLens}\label{tunedlens_appendix}
TunedLens \cite{belrose2023eliciting} is an extention of LogitLens \cite{nostalgebraist2020logitlens}, which projects the transformer layer residual streams into distributions over the model vocabulary. LogitLens applied to a hidden state, $h$, at layer $l$ is defined as
\[
\text{LogitLens}_l(h_l)=\text{LayerNorm}(h_l)W_U,
\]
where $\text{LayerNorm}$ is the output normalization layer of the transformer, and $W_U$ is the transformer unembedding matrix. Using the final layer unembedding matrix, LogitLens generates a token prediction given the residual stream state at the current layer. Applied to each transformer layer, the evolution of predictions during the forward pass can be observed.

However, it has been shown that the LogitLens is a biased estimator of the final logit distribution, and that it is also susceptible to residual stream covariance drift \cite{belrose2023eliciting}. Hence, the predictions it computes are not trustworthy approximations of the contents of the residual stream, and it cannot perform well across different transformer architectures. TunedLens improves on LogitLens by decreasing the effect of representation drift and bias using a learnable affine projection. This allows TunedLens to map the hidden state of a given layer to a similar space as the hidden state of the final transformer layer. Formally, the TunedLens at layer $l$ is computed as:
\[
\text{TunedLens}_l(h_l) = \text{LogitLens}_l(A_lh_l+b_l)
\]
where $A_l$ and $b_l$ are the affine projection weights and biases for layer $l$. Each TunedLens probe is trained to minimize the KL divergence between current layer token distribution and the transformer output logits.

\begin{figure}
    \centering
    \includegraphics[width=\textwidth]{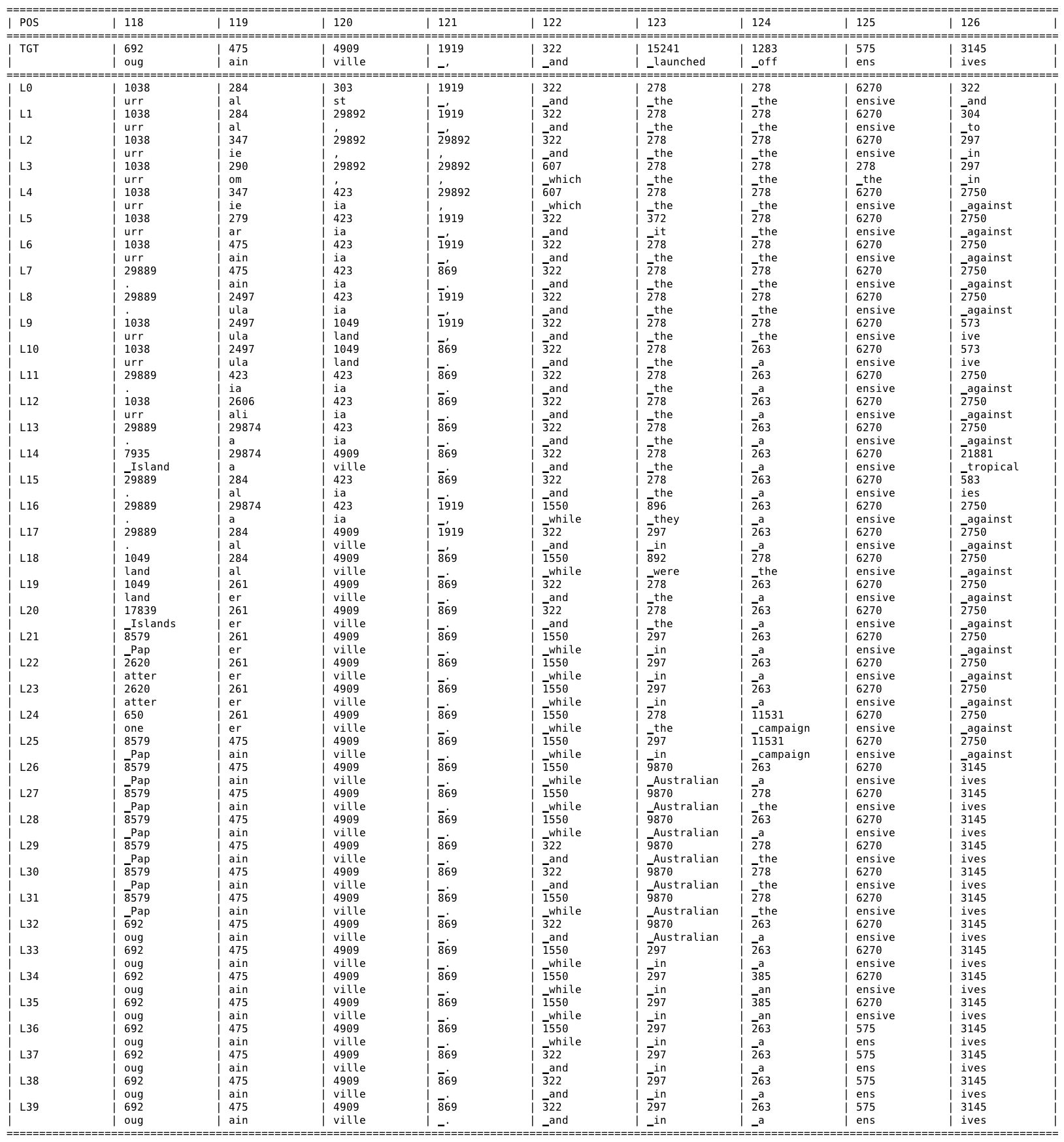}
    \caption{\textbf{Toy TunedLens + \FlexModel{}} example inference. The \texttt{POS} label specifies the token position in the context, the \texttt{TGT} label specifies the model output argmax token, and each row corresponds to the TunedLens probe argmax token at layer \texttt{L}.}
    \label{fig:tunedlens-inf}
\end{figure}

As a toy demonstration of TunedLens using \FlexModel{}, a simplified version of the training procedure is run on LLaMA-2-13B. The model is wrapped with \FlexModel{}, and \HookFunction{}s are applied to each transformer layer residual stream. Additionally, \FlexModel{} provides functions for retrieving unsharded parameters in a similar fashion to activation retrieval using \HookFunction{}s. This is used to collect the unsharded RMSNorm and output embedding weights to be applied to each TunedLens probe. Using the \FlexModel{} API, all of the interactions required between the TunedLens model and the wrapped LLaMA-2-13B model are condensed to a small code-block, see Figure \ref{fig:tunedlens_code}. The experiment is run using the WikiText-103 dataset \cite{merity2016pointer} and a short maximum sequence length of $128$ tokens to reduce the resources and time required for the experiment. See Figure \ref{fig:tunedlens-inf} for an example inference. Even in this toy example, we observe the predictions made by each layer being steadily refined during the forward pass.

\begin{figure}[ht!]
    \begin{minted}[breaklines, breakindent=7pt, breaksymbolleft=]{python}
    class TunedLens(nn.Module):
        ...
        # Setup Flex Model wrapper.
        self.act_dict = {}
        self.flex_model = FlexModel(
            base_model,
            self.act_dict,
            tp_size, pp_size, dp_size,
        )
        # Hook into residual stream states.
        for layer_name in residual_stream_layers:
            self.flex_model.register_hook_function(
                HookFunction(
                    layer_name,
                    expected_shape=(None, None, hidden_dim),
                )
            )
        # Retrieve unsharded weights.
        unembed_weight = self.flex_model.get_module_parameter(
            "output.weight",
            (vocab_size, hidden_dim),
        )
        norm_weight = self.flex_model.get_module_parameter(
            "norm.weight",
            (hidden_dim,),
        )
        # Init Norm and Unembed TunedLens layers using weights.
        ...
            
    \end{minted}
    \caption{TunedLens initialization example using \FlexModel{} and \HookFunction{}s. \FlexModel{} is used to wrap the base model, and \HookFunction{}s are placed at each layer to retrieve the unsharded residual stream activations. The weights for the normalization and unembedding layers are also fetched and unsharded by \FlexModel{} to generate the residual stream logit distributions.}
    \label{fig:tunedlens_code}
\end{figure}

\section{Additional Induction Head Search Results} \label{llama_induction_appendix}

In this section, the results of an induction head search on LLaMA-2-13B are reported. As done for LLaMA-2-70B, the model is distributed using FSDP over for A100 GPUs. Several induction heads are identified in the early layers of the model.

\begin{figure}
     \centering
     \begin{subfigure}[b]{0.49\textwidth}
         \centering
         \includegraphics[width=\textwidth]{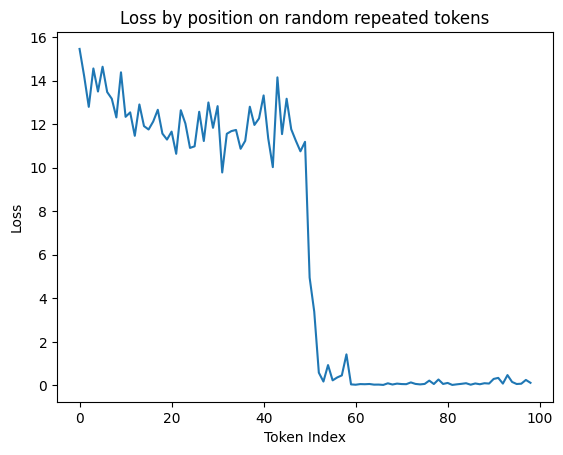}
         \caption{}
         \label{fig:tloss_13b}
     \end{subfigure}
     \begin{subfigure}[b]{0.46\textwidth}
         \centering
         \includegraphics[width=\textwidth]{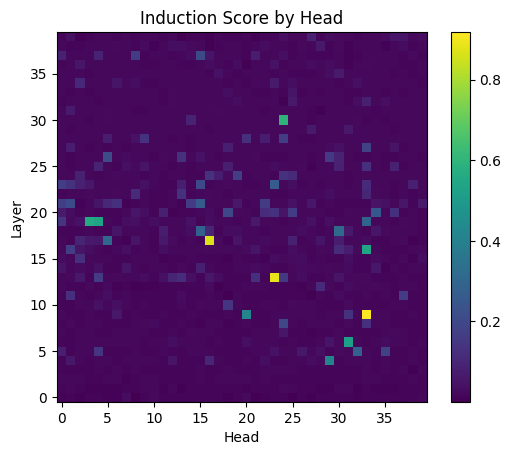}
         \caption{}
         \label{fig:iscore_13b}
     \end{subfigure}
    \caption{Empirical verification of the presence of induction heads within LLaMA-2-13B using a repeated randomly generated sequence as input. The measured loss per token is shown in (\subref{fig:tloss_13b}) and the location of the induction heads is shown in (\subref{fig:iscore_13b}), indicated by the high induction score values.}
    \label{13b}
\end{figure}




\end{document}